\documentclass[journal]{elsarticle}

\usepackage[hidelinks]{hyperref}
\usepackage{hyperref}
\usepackage{subcaption}
\usepackage{epstopdf}
\usepackage{graphicx}
\usepackage{amsfonts}
\usepackage{enumerate}
\usepackage{varioref}
\usepackage{longtable}
\usepackage{graphicx}
\usepackage{amssymb}
\usepackage{chngpage}
\usepackage{multirow}
\usepackage{multirow}
\usepackage{amsthm}
\usepackage{amsmath}
\usepackage{caption}
\usepackage{nccmath}
\usepackage[table]{xcolor}
\usepackage{algorithm}
\usepackage{algpseudocode}
\usepackage{lineno}
\hypersetup{colorlinks=true,citecolor=green}
\DeclareMathOperator*{\argmin}{argmin}

\newdimen{\algindent}
\algnewcommand\LeftComment[2]{\hspace{#1\algindent}$\triangleright$ {#2} \hfill}

\begin{document}

\title{Simultaneous Joint and Object Trajectory Templates for Human Activity Recognition from $3$-D Data}
\author[1]{Saeed Ghodsi}
\ead{saeed.ghodsi@ee.sharif.edu}
\author[1]{Hoda Mohammadzade\corref{cor}}
\ead{hoda@sharif.edu}
\author[1]{Erfan Korki}
\ead{erfan.korki@alum.sharif.edu}
\cortext[cor]{Corresponding Author}
\address[1]{\footnotesize Department of Electrical Engineering, Sharif University of Technology, Tehran, Iran.}

\begin{frontmatter}

\begin{abstract}
The availability of low-cost range sensors and the development of relatively robust algorithms for the extraction of skeleton joint locations have inspired many researchers to develop human activity recognition methods using the $3$-D data. In this paper, an effective method for the recognition of human activities from the normalized joint trajectories is proposed. We represent the actions as multidimensional signals and introduce a novel method for generating action templates by averaging the samples in a "dynamic time" sense. Then in order to deal with the variations in the speed and style of performing actions, we warp the samples to the action templates by an efficient algorithm and employ wavelet filters to extract meaningful spatiotemporal features. The proposed method is also capable of modeling the human-object interactions, by performing the template generation and temporal warping procedure via the joint and object trajectories simultaneously. The experimental evaluation on several challenging datasets demonstrates the effectiveness of our method compared to the state-of-the-arts.
\end{abstract}


\end{frontmatter}

\section{\textbf{Introduction}}
Human activity recognition (HAR) is one of the most important research areas in computer vision. In HAR, the purpose is to utilize human movement data (e.g. an RGB video), in order to identify performed activities. Based on the complexity, human activities are usually classified into four categories: gestures, actions, interactions, and group activities \cite{aggarwal2011human}. Recognition of the human activities enables a broad range of applications from automated surveillance systems, patient and elderly monitoring systems, and personal assistive robotics to a variety of systems that involve human-computer interaction \cite{lun2015survey}. In this paper, we concentrate on the recognition of human actions as the combination of elementary body part movements.

Here we divide activity recognition challenges, into two major types. Low-level challenges are related to our data gathering method and environmental conditions. For example, view angle, size, and illumination variations, as well as occlusion, cluttering, and shadows are in this group. On the other side, high-level challenges are caused by the nature of the actions. It should be considered that individuals can perform the same action with different styles and different speeds. Even one person, depending on the situation, can perform a specific action in different ways.

Development of activity recognition methods began in the early '80s. Till recent years, research in this area was mainly focused on the recognition via 2-D video cameras. The recent availability of depth sensors with admissible precision and reasonable cost and size, motivated the computer vision community to conduct more research on the $3$-D based action recognition. Aggarwal et al. \cite{aggarwal2011human} divided the $3$-D data acquisition methods into three categories: marker-based motion capture systems, multi-view stereo images, and range sensors. The utilization of range sensors significantly alleviates the low-level challenges explained previously. Based on the extracted features from the $3$-D data, Aggarwal et al. \cite{aggarwal2014human} classified recognition methods into five groups: features from $3$-D silhouettes, features from skeletal joint locations, local spatiotemporal features, local occupancy patterns, and $3$-D scene flow features.

\begin{figure}
\includegraphics[width=\linewidth]{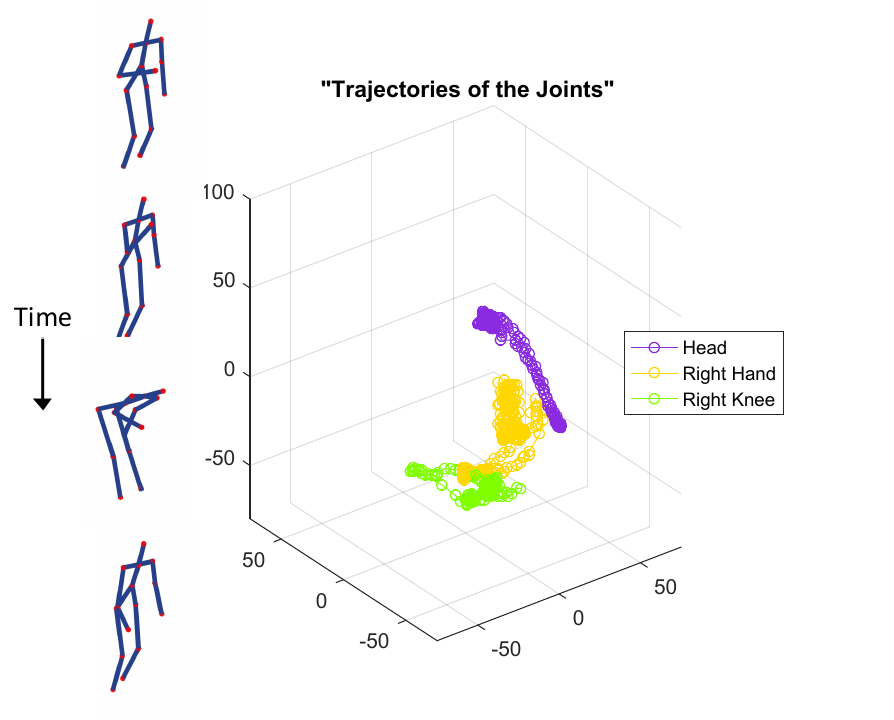}
\caption{Joint trajectories of the "Rinsing Mouth" action from the \textquotedblleft CAD-60\textquotedblright dataset.} \label{fig:1}
\end{figure}

In this paper, we propose an activity recognition system, using the $3$-D location of joints and objects, extracted from the depth image sequences. We represent the human action as a set of trajectories, corresponding to the skeleton joints locations along time (Fig. \ref{fig:1}). To make our method robust against the different styles of performing actions, we transform the joints to a human-centric coordinate system, in which, the trajectories are extracted. In this representation, human object interactions can also be modeled similarly by relative object trajectories. Then we propose a novel algorithm for the construction of template joint and object trajectories to effectively represent the actions. We also present a template-based sequence warping approach to deal with the effect of varying style, speed, and acceleration of the subjects. To consider the locality in both time and frequency domains, wavelet features are extracted from the trajectory signals. The classification results demonstrate that our proposed method is efficient and gives comparable results to the state-of-the-art approaches on several datasets.

The remainder of this paper is organized as follows. An overview of the most related methods is presented in section \ref{related_work_section}. In section \ref{methodology_section}, we first describe the preprocessing of the skeleton data, and motion representation steps. Then the template generation and temporal warping algorithms are introduced, and finally, the feature extraction and classification strategies are illustrated. Section \ref{experiments_section} is the discussion and comparison of the experimental results of our algorithm on multiple datasets, and section \ref{conclusion_section} is the conclusion of the paper.

\section{\textbf{Related Work}}\label{related_work_section}
In this section, a concise review of skeleton-based activity recognition methods is presented. More details are provided in \cite{han2016space}, \cite{presti20163d}, and \cite{ye2013survey}. We also refer the interested readers to \cite{aggarwal2011human} and \cite{weinland2011survey} for a review on RGB video-based approaches and \cite{aggarwal2014human}, \cite{ye2013survey}, and \cite{chen2013survey} for depth map-based approaches. In the following, we will review different works, from the perspective of skeletal joints representation, and the temporal modeling methodology.

In the literature, different representations are proposed for human activities. Many methods directly use the raw joint positions. Considering the location of joints as random variables, Hussein et al. \cite{hussein2013human} formed vectors to describe the actions, and then computed the covariance matrices of the vectors, to form the feature vector. Inspired by the idea of temporal pyramids, multiple covariance matrices are calculated over different windows of frames, to maintain the temporal order of the actions. Zanfir et al. \cite{zanfir2013moving} proposed the moving pose descriptor, which included the information of positions, as well as, speed and acceleration of the joints. In \cite{yang2014effective} the combination of feature vectors from the raw joint locations, pairwise distances between joints, and the motion of the joints are extracted and normalized. Then the Eigenjoints are generated by applying the Principle Components Analysis. To improve the recognition accuracy, Zhu et al. \cite{zhu2013fusing} tried to fuse skeletal joints features with spatiotemporal features. The authors used well-known image feature point detectors and descriptors, such as Histogram of Gradients (HOG), and Speeded-up Robust Features (SURF), to extract features from the depth maps. Skeletal features are extracted in the same way as \cite{yang2014effective}, and after quantization with the k-means algorithm, histograms of features are fused together using the Random Forest classifier. Representation of the actions is sometimes performed by modeling the geometric relationships between the body parts. Vemulapalli et al. \cite{vemulapalli2016r3dg} introduced the so-called R3DG features, i.e. a family of skeleton representations. They model the human skeleton via $3$-D body transformations and represent human actions as R3DG curves.

Instead of using handcrafted features, deep learning methods attempt to explain the raw data in an automatic manner. Du et al. \cite{du2015hierarchical} divided human skeleton into five distinct body parts and utilized a hierarchical structure of Bidirectional Recurrent Neural Networks (BRNNs) to represent the actions. In the first layer of the network, raw positions of the body parts joints were fed into the corresponding RNNs. Then the inputs of each layer were formed by a combination of the outputs of the previous layer. A fully connected layer with softmax activation was used to perform the classification. Similarly, Zhu et al. \cite{zhu2016co} proposed a three layered Long Short-Term Memory (LSTM) structure to learn human representations from the joint trajectories. Both the spatial and temporal information of the skeletal joints were utilized in \cite{liu2016spatio} to train a spatiotemporal LSTM network. A Trust Gate was also proposed, to deal with the noise due to the joint location extraction. Wu and Shao \cite{wu2014leveraging} extracted features from the skeleton joint locations and then adopted deep belief networks to estimate the emission probabilities in Hidden Markov Models (HMMs).

Trajectory-based methods, consider an action, as a set of multiple time series representing the location of different joints over time, and extract features from the trajectories. Gupta et al. \cite{gupta20143d} introduced a motion-based descriptor to compare the Mocap data with the trajectories extracted from videos directly and generates multiple motion projections as their feature. Wei et al. \cite{wei2013concurrent} applied the wavelet transform and extracted features from the trajectories to address the problem of concurrent action detection. The self-similarity based descriptor, proposed by Junejo et all. \cite{junejo2011view}, is an encoding mechanism for the temporal shapes of human actions observed in the videos. Experimental evaluations have shown the stability of this representation under view changes. Many methods transform the trajectories in the Euclidean space into curves in a manifold. Devanne et al. \cite{devanne20153} proposed transforming motion trajectories into a Riemannian manifold and performing the classification using the Nearest Neighbor methods. In \cite{slama2015accurate} trajectories are represented as points in the Grassmann manifold. Then the learning procedure is performed by the calculation of Control Tangents for the action clusters. Amor et al. \cite{amor2016action} modeled trajectories on Kendall’s shape manifold and introduced a new framework for the temporal alignment of the trajectories to handle the challenge of execution rate variance of the actions. Gong and Medioni \cite{gong2011dynamic} proposed a Spatio-Temporal Manifold (STM) to model the human joint trajectories over time. They also adapted the idea of Dynamic Time Warping to provide an algorithm for the alignment of time series under the STM model, called Dynamic Manifold Warping (DMW).

Another group of methods, try to learn dictionaries of code-words, extracted from the skeleton \cite{chaudhry2013bio}, \cite{wu2015watch}. In \cite{zhu2016human} multi-layer codebooks of key poses and atomic motions were learned using the relative orientations of body limbs. Then the action patterns were represented via the codebooks of each action, and a pattern matching algorithm was proposed to recognize the actions. Xia et al. \cite{xia2012view} calculated Histograms of $3$-D Joint locations (HOJ3D), by partitioning the space around the body of the subject to a total number of 84 bins and counting the number of joints falling in each bin. The resulting histogram represents the posture of the body. The K-means clustering algorithm is then utilized for quantization and generation of the posture vocabulary. Feeding the time domain sequences of the code-words into Hidden Markov Models (HMMs), yields statistical models representing the whole actions. Similarly, Wang et al. \cite{wang2013approach} grouped skeletal joints into five body parts and generated spatial and temporal dictionaries to represent the actions, using the K-means algorithm. Combining the group sparsity and geometry constraints, Luo et al. \cite{luo2013group} proposed a sparse coding algorithm, to learn the dictionary, based on the relative joint locations. 

Some trajectory-based approaches employ the idea of dictionary learning in the form of action templates. Muller and Roder \cite{muller2006motion} introduced the concept of motion templates to represent the actions, and then performed the recognition by a Nearest Neighbor classifier. Pairwise distances of the skeleton joints were used in \cite{zhao2013online} to learn a dictionary of motion templates. Then the Structure Streaming Skeleton (SSS) features are computed and a sparse coding approach is used for the gesture modeling. Vemulapalli et al. \cite{vemulapalli2014human} introduced a representation for the motion trajectories, as curves in the Lie Group $SE(3) \times\cdots\times SE(3)$. To simplify the task of classification of the curves and be able to apply standard temporal modeling methods, they mapped the curves into the corresponding Lie Algebra. Then nominal curves for the actions were computed, and all the samples were warped to the curves. Following Wang et al. \cite{wang2012mining}, the Fourier Temporal Pyramid (FTP) was applied, and a set of Support Vector Machines (SVMs) were adopted to perform the classification.

Due to the different discrimination power of the body joints for the recognition of actions, many methods tried to mine for the most informative joints. The proposed algorithm by Chaaraoui et al. \cite{chaaraoui2014evolutionary} attempts to find a subset of joints, which performs the recognition task better than all joints. Dynamic Time Warping (DTW) distance of the joint location trajectories was used in \cite{reyes2011featureweighting} to measure the similarity of the action sequences. To determine the impact of each joint on the total distance function, the weighting values of joints were computed by calculating the amount of similarity of the joints trajectories in each class and dissimilarities of the trajectories between distinct classes. By determining the most informative subset of the joints for each specific action class in consecutive time segments, and then concatenating them, Ofli et al. \cite{ofli2014sequence} proposed a novel representation of the actions. Pairwise distances between the joints as well as Local Occupancy Patterns (LOP) around the joints were employed as features in \cite{wang2012mining}. Then Fourier Temporal Pyramid (FTP) was applied to make the representation robust against the temporal misalignment and noise. Moreover, an actionlet-based approach was introduced to mine for the most discriminative combination of the joints using the multiple kernel learning method.

In some activities, the human object interactions play an important role. In the literature, many methods have been proposed to model the human object interaction. Inspired by the idea of dividing a high-level human activity into smaller atomic actions, Wei et al. \cite{wei2013modeling} introduced a hierarchical graph to represent the human pose in the $3$-D space, and the motions through 1-D time. They defined an energy function, interpreted by the graph, which consists of two terms. The spatial term, includes the pose model, object model and the geometric relations between the skeleton and objects, and the temporal term includes atomic events transition and object motions. Similarly, Koppula et al. \cite{koppula2013learning} aimed at jointly learning the human activities and object affordances, by defining a Markov Random Field (MRF) with two kinds of nodes, corresponding to the objects and the sub-activities. The motion and position of the objects were fed to the object node as the feature vector, and the human object interactions were modeled by the graph edges. In contrast with these works, a single layered approach was proposed in Tayyub et al. \cite{tayyub2014qualitative}, to model the human object interactions, regardless of the object type. They extract qualitative and quantitative features from the objects, in the spatial and temporal domains, and apply a feature selection technique to recognize the actions efficiently. Their experiments suggested that the spatial features, i.e. the relations between the different objects in the $3$-D space, have a major impact on the discrimination between distinct activities.

\section{\textbf{Methodology}}\label{methodology_section}
In this section, first, we explain the preprocessing of the raw $3$-D data and action representation strategy. We then explain the action template generation and temporal warping steps, followed by the description of the feature generation and classification methods.

\subsection{\textbf{{Action Representation}}}
\begin{figure}
\includegraphics[width=\linewidth]{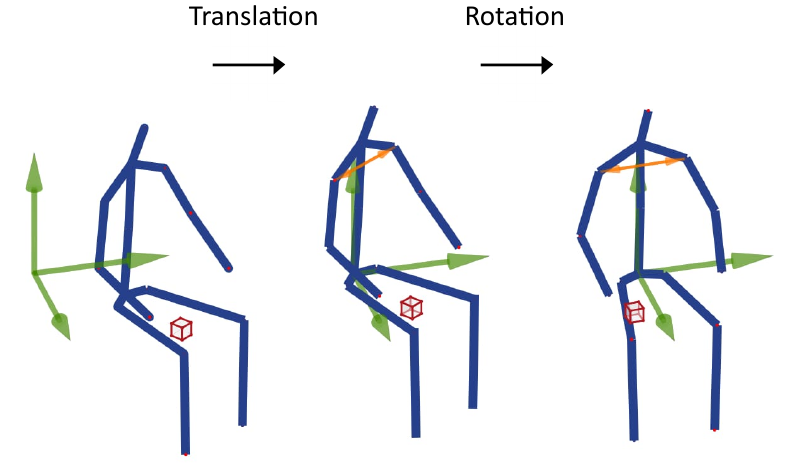}
\caption{An illustration of the alignment procedure.} \label{fig:3}
\end{figure}

In this paper, we use a trajectory-based action representation. We model an action sample, as a set of multiple time series, each representing the variations of one coordinate of the position of one skeleton joint over time. If the actions include human-object interactions, we extract the $3$-D positions of the objects and form the object trajectories. Then similar to the body joints, the object trajectories are also utilized for the action representation. Preprocessing of the raw data is usually performed to cope with the low-level challenges mentioned previously. To eliminate the effect of different positions of the subject with respect to the camera and make our method robust against the viewpoint variance, we perform a skeleton alignment procedure in each frame. For this purpose, we transform the $3$-D positions of the skeleton joints, from the camera coordinates to a person-centric system by moving the hip joint of the subject to the origin and rotating the skeleton along the $z$-axis to a predefined orientation. This geometric transformation is identical to first calculating the connecting vectors from the skeleton joints and the tracked objects to the hip joint, and then applying the same rotation to all the resulting vectors. The same translation and rotation are applied on the different skeleton joints. Some differences in the style of performing actions, such as different directions in the "walking" action, or minor body movements while "drinking water" action, will be handled by performing the aforementioned geometric alignment on each frame. This alignment procedure, which is illustrated in Fig. \ref{fig:3}, is similarly applied on all the tracked objects. More specifically, for each object, the locations of the objects $2$-D bounding boxes in the RGB images are extracted by means of an off-the-shelf object detection and tracking algorithm. Then using the corresponding depth map images and the Kinect's camera calibration parameters, the real world $3$-D coordinates of the object are determined along time. The extracted trajectories of the objects are used in the alignment procedure.

Let $\mathcal{J}$ and $\mathcal{O}$ be the number of tracked skeleton joints, and the maximum number of manipulated objects between the actions, respectively. Suppose $\mathfrak{S}^{(i,j)}$ be the $j$-th sample of the $i$-th action class. So the sample can be represented by the set of $\mathfrak{S}^{(i,j)} = \{\mathfrak{S}^{(i,j)}_k, k=1,2,\cdots,\mathcal{K}\}$, where $\mathcal{K}=(\mathcal{J}+\mathcal{O})\times3$ denotes the number of time series, and each $\mathfrak{S}^{(i,j)}_k$ is a single time series, corresponding to the variations of the $x$, $y$, and $z$ coordinates of one skeleton joint or tracked object in the time domain. Since the different number of objects can be present in different actions, we make the number of objects equal by placing some extra objects in the hip joint location of the subject, when needed. For example, if the actions involve at most five object manipulations, and an action has three objects, we put two extra objects in the hip joint location to make the number of time series equal. Hereafter, we consider the whole set of time series, representing an action sample, as a multidimensional signal, and name each single time series as a sub-signal. Note that the trajectories of the joints and objects are formed in the person-centric coordinates system. Then we apply a Savitzky-Golay smoothing filter \cite{savitzky1964smoothing} on the sub-signals to reduce the effect of noise, due to the depth image extraction by the Kinect sensor and the minor errors of the joints and objects position estimation. A median filter is also utilized to remove the joint position spikes.

\subsection{\textbf{Temporal Warping}}
One major issue in the action classification is the varying length and velocity of actions due to the different styles of performing actions. In the trajectory-based methods, usually Dynamic Time Warping (DTW) is utilized to deal with the temporal variations. DTW is an algorithm to find the optimal match between two given time series. Warping a sequence with another one means determining the non-linear correspondence between the time indices of the sequences, which best represents the shape similarity of them. DTW attempts to handle the deformations of the sequences in the time domain, by assigning each index in one sequence, to zero , one or more indices in the other sequence depending on the similarity between them. The output of the algorithm is the distance between the two sequences, which is defined to be the sum of the squared distances between the value of the signals at their matched indices, and also the ordered pair of the matched indices.

DTW can be employed to classify the sequences. As an example, a simple Nearest Neighbor classifier with the DTW distance measure can be adopted to determine the most similar pre-labeled action sequence to the input test sequence. Although having enough training samples, this method yields relatively good results, but the DTW algorithm is very slow in practice, even when implemented with dynamic programming techniques. Therefore comparing an input test sample with a lot of pre-labeled samples with DTW is very time-consuming and probably not appropriate for many real world applications. To cope with this challenge, we propose to warp the samples of each action, with a corresponding pre-trained action template. We first create one template for each action class in the training phase, and then in the test phase, we will use the DTW to warp the input sample merely with the templates. Thus, instead of performing DTW with many samples for each action class, we just perform the calculation with one template per action, making it much simpler.

Before explaining the template generation algorithm, we define the "mean-sample" of an action class. Let $\mathfrak{S}^{(i,j)}$, $j=1,2,\cdots,\mathcal{N}^i$ be the set of samples of the $i$-th action. The "mean-sample" of an action is a set of the $\mathfrak{S}^{(i,j)}_k$ sub-signals, which are most similar to the other corresponding sub-signals of this class. We find this sample by a method similar to the one proposed by Gupta and Bhavsar \cite{gupta2016scale}.
\begin{algorithm}{}
\caption{Mean-Sample Search Algorithm}\label{alg1}
\begin{algorithmic}[1]
\State Given $\mathfrak{S}^{(i,j)}$, $\forall i,j\,$
\For {$i = 1,\cdots,\mathcal{C}$}
	\For {$k = 1,\cdots,\mathcal{K}$}
		\For {$j = 1,\cdots,\mathcal{N}^i$}
			\Statex \LeftComment{3} {Sum up the $DTW$ distances:}
			\State $\mathcal{\zeta}^j$ $\gets$ $\sum_{j'=1}^{\mathcal{N}^i} DTW{(\mathfrak{S}^{(i,j)}_k,\mathfrak{S}^{(i,j')}_k)}$
		\EndFor
		\State $\hat{j} $ $\gets$ $\argmin_j \{\mathcal{\zeta}^j\}$
		\State $\mathcal{M}^{(i)}_k$ $\gets$ $\mathfrak{S}^{(i,\hat{j})}_k$
	\EndFor
\EndFor
\State \Return $\mathcal{M}^{(i)}$, $\forall i\,$
\end{algorithmic}
\end{algorithm}
The method for finding the mean sample is described in Alg. \ref{alg1}, where $\mathcal{C}$, and $\mathcal{N}^i$ are the number of action classes, and the number of training samples for the $i$-th class respectively. In Alg. \ref{alg1}, the distance of the $\mathfrak{S}^{(i,j)}_k$ and $\mathfrak{S}^{(i,j')}_k$ sub-signals, is defined as the DTW distance of the two time series. The total distance value for each sub-signal of each training sample is defined as the summation of the distances from this sample to the others. The "mean-samples" are then found by minimizing the total distance values of the samples within each class. Since we calculate the sub-signals of the "mean-samples" separately, these sub-signals might come from different samples, and therefore they might have different lengths. Experimental results demonstrate the superiority of this algorithm over other algorithms in which one of the samples are chosen as the mean sample directly.

\begin{figure}[htbp]
\centering
\begin{subfigure}[b]{0.45\textwidth}
\includegraphics[width=\linewidth]{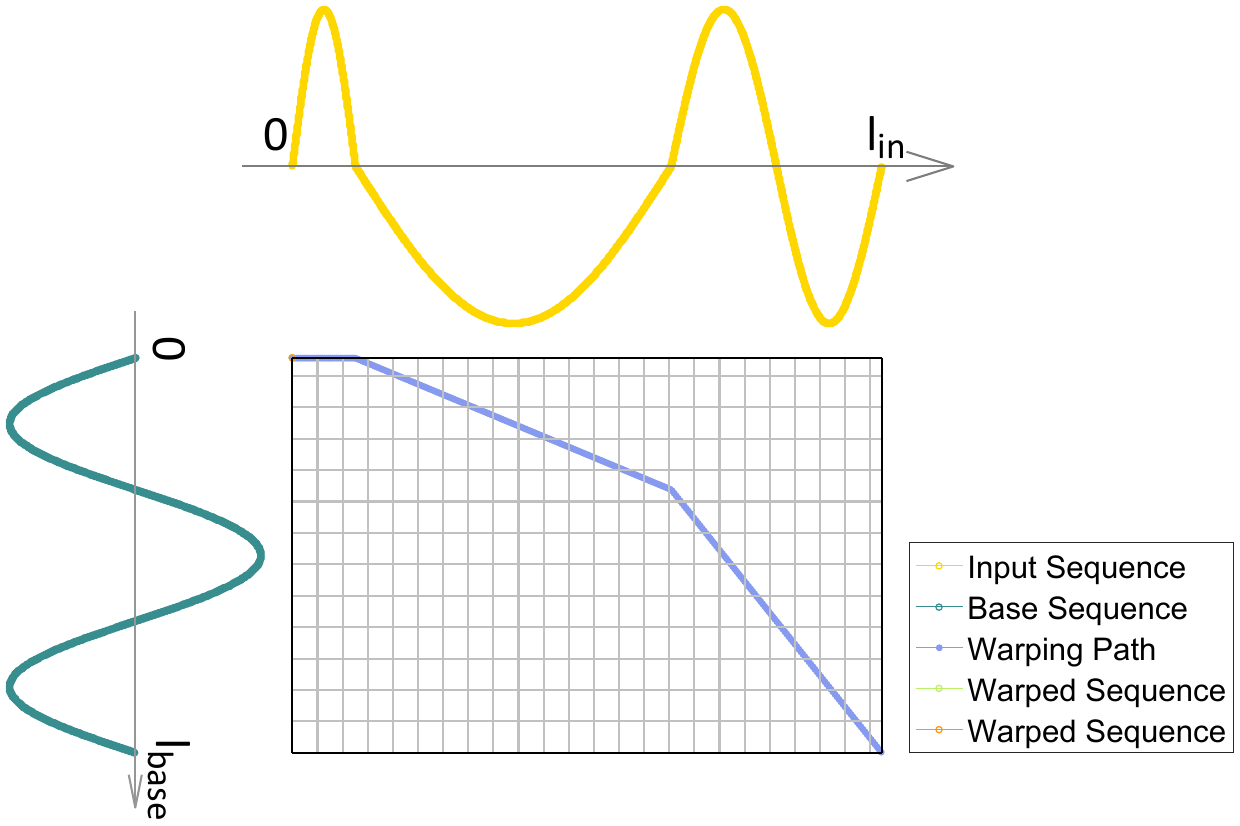}
\caption{Warping Path} \label{fig:4a}
\end{subfigure}
\hspace*{\fill} 
\begin{subfigure}[b]{0.45\textwidth}
\includegraphics[width=\linewidth]{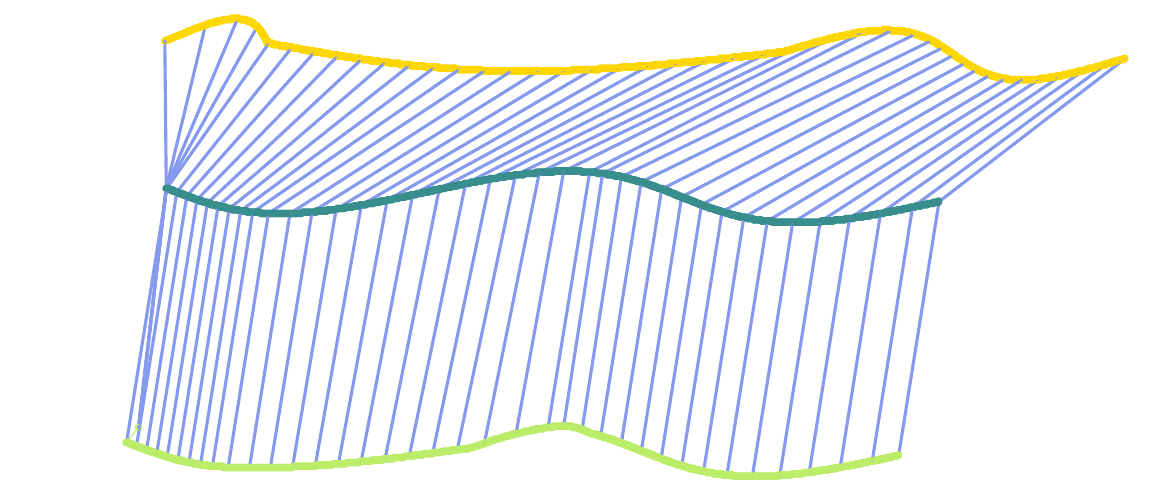}
\caption{Fine Warping} \label{fig:4b}
\end{subfigure}
\par\bigskip 
\begin{subfigure}[b]{0.45\textwidth}
\includegraphics[width=\linewidth]{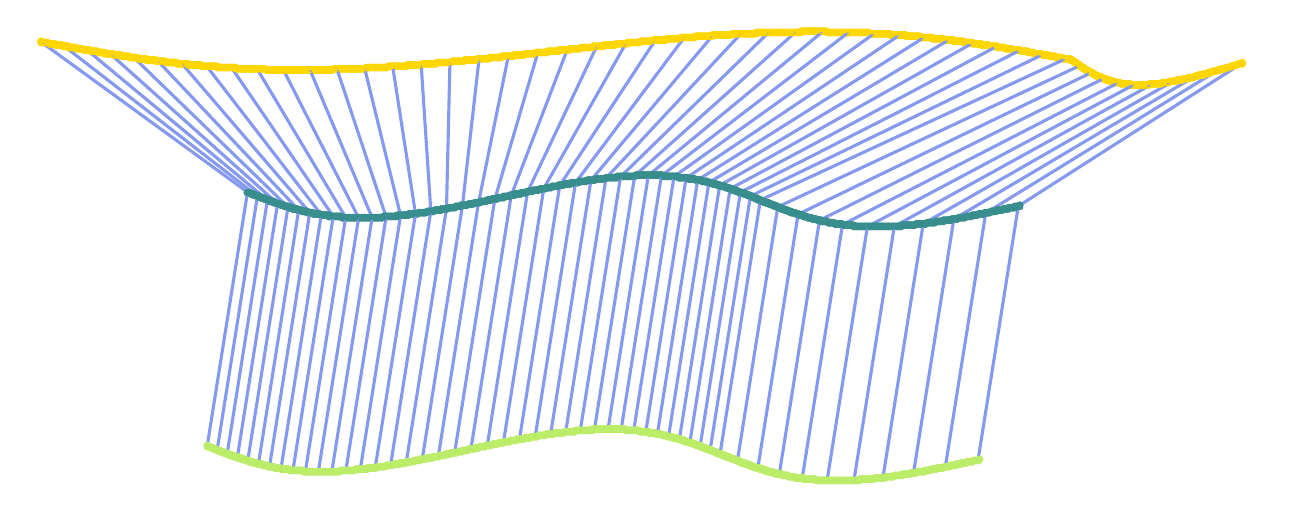}
\caption{Ideal Warping} \label{fig:4c}
\end{subfigure}
\hspace*{\fill} 
\begin{subfigure}[b]{0.45\textwidth}
\includegraphics[width=\linewidth]{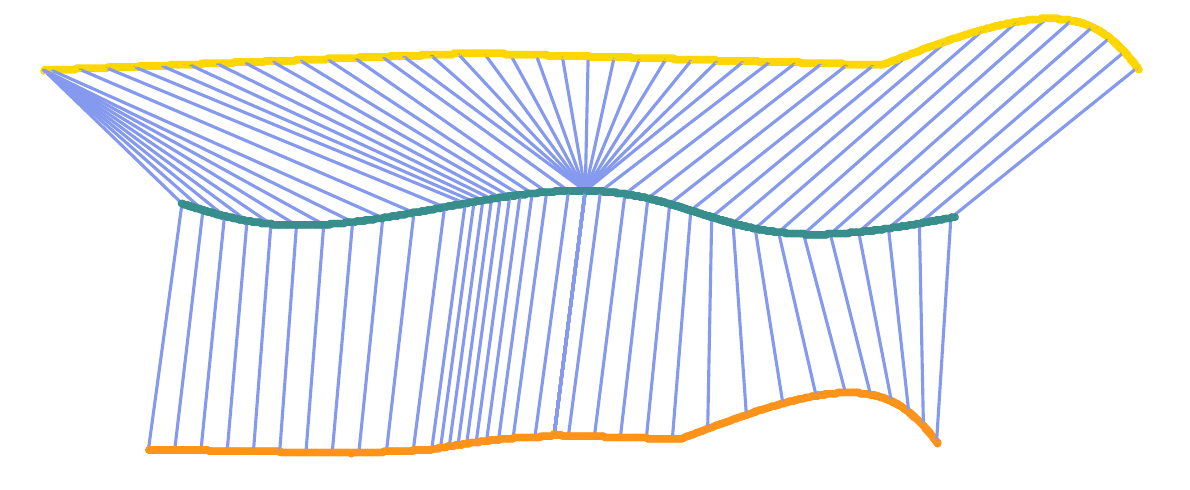}
\caption{Bad Warping} \label{fig:4d}
\end{subfigure}
\caption{Examples of the sequence warping procedure.} \label{fig:4}
\end{figure}

Next, we will use the "mean-samples", to achieve better representations of the action. First, we explain the algorithm for warping of a multidimensional signal with another one (Alg. \ref{alg2}). Let $\mathfrak{S}$ and $\mathfrak{S}'$ be two arbitrary action samples. To warp $\mathfrak{S}$ with $\mathfrak{S}'$, we perform the DTW between each pair of the corresponding sub-signals, $\mathfrak{S}_k$ and $\mathfrak{S}'_k$, $k=1,2,\cdots,\mathcal{K}$, and compute the optimal matching paths. Then for each $\mathfrak{S}'_k$, iterating on the indices of this time series, the value of the matched index in $\mathfrak{S}_k$ is used as the warped value of the corresponding index. If there are multiple indices assigned to one index, we'll average the values to obtain the correct warped value. It is also possible that some indices of $\mathfrak{S}_k$, wouldn't have any matching on the other side. In this case, we linearly interpolate the sequence for the missing value. All of the sub-signals are warped in this way with the corresponding sub-signals in the base multidimensional signal. At the end of this procedure, we will have the new set of sub-signals, maintaining their overall shape, while matching in the length with the base sub-signals. Some examples of sequence warping are illustrated in Fig. \ref{fig:4}.

\begin{algorithm}{}
\caption{Warping Algorithm}\label{alg2}
\begin{algorithmic}[1]
\Procedure{Warp}{$\mathfrak{S}$, $\mathfrak{S}'$}
\For {$k = 1,\cdots,\mathcal{K}$}
	\Statex \LeftComment{2} {$DTW$ returns the distance and warping paths}
	\State $[\mathcal{\zeta},\mathcal{P},\mathcal{P}']$ $\gets$ $DTW({\mathfrak{S}_k,\mathfrak{S}'_k})$
	\State $i$ $\gets$ $1$
	\State $\mathcal{L}$ $\gets$ $Len(\mathfrak{S}'_k)$
	\For {$l = 1,\cdots,\mathcal{L}$}
		\State $\sigma$ $\gets$ $0$ $,$
		$n$ $\gets$ $0$
		\While {$\mathcal{P}'(i)=l$}
			\State $\sigma$ $\gets$ $\sigma + \mathfrak{S}_k[\mathcal{P}(i)]$
			\State $n$ $\gets$ $n + 1$ $,$
			$i$ $\gets$ $i + 1$
		\EndWhile
		\If {$n \geq 1$} $\mathcal{W}_k[l]$ $\gets$ $\dfrac{\sigma}{n}$
		\Else \thinspace $\mathcal{W}_k[l]$ $\gets$ $linear \, interpolation$
		\EndIf
	\EndFor
\EndFor
\State \Return $\mathcal{W}$
\EndProcedure
\end{algorithmic}
\end{algorithm}

\begin{figure}
\includegraphics[width=\linewidth]{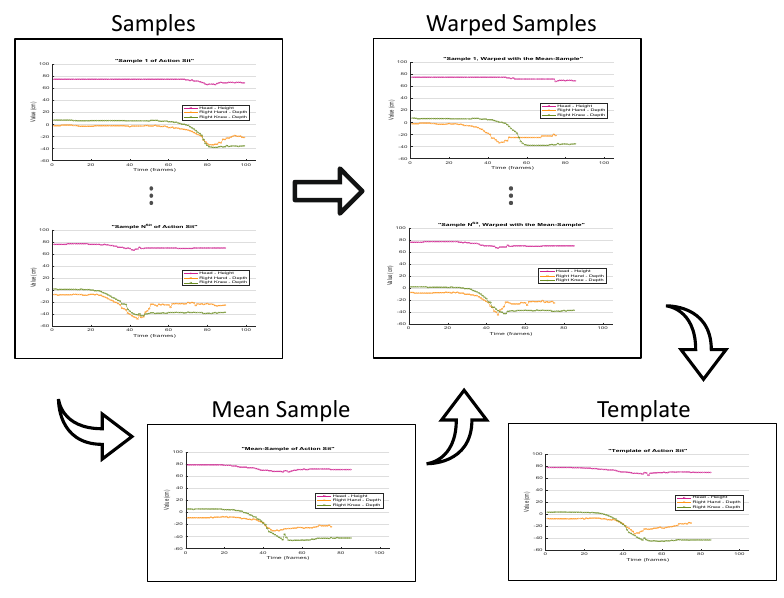}
\caption{Illustration of the template generation algorithm for action "Sit" from the \textquotedblleft TST Fall Detection\textquotedblright dataset.} \label{fig:5}
\end{figure}

Now, for each action class, we create a new multidimensional signal, called "action template", as described in Alg. \ref{alg3}. Although templates are being generated on the basis of the corresponding "mean-samples", but, utilizing a kind of averaging method, we attempt to make them more similar to the training samples of the action. To create the template, we warp all the training samples of the class, with the "mean-sample", as explained above. Then, since all the resulting samples are the same length, we can perform a simple averaging on each index of each sub-signal, to obtain the template. An example of the template generation algorithm is presented in Fig. \ref{fig:5}.

\begin{algorithm}{}
\caption{Template Generation Algorithm}\label{alg3}
\begin{algorithmic}[1]
\For {$i = 1,\cdots,\mathcal{C}$}
	\For {$j = 1,\cdots,\mathcal{N}^i$}
		\State $\mathfrak{S}'^{(i,j)}$ $\gets$ $WARP(\mathfrak{S}^{(i,j)},\mathcal{M}^{(i)})$
	\EndFor
	\For {$k = 1,\cdots,\mathcal{K}$}
		\State $\mathcal{L}$ $\gets$ $Len(\mathcal{M}^{(i)}_k)$
		\For {$l = 1,\cdots,\mathcal{L}$}
			\State $\mathcal{T}^i_k[l]$ $\gets$ $\dfrac{\sum_{j=1}^{\mathcal{N}^i} \mathfrak{S}'^{(i,j)}_k[l]}{\mathcal{N}^i}$
		\EndFor
	\EndFor
\EndFor
\State \Return $\mathcal{T}^i$
\end{algorithmic}
\end{algorithm}

Finally, the pre-trained templates are used to warp the samples, of both training and testing sets. We warp each sample, regardless of its class, with the templates of all actions. So if we have $\mathcal{C}$ actions in total, we will have $\mathcal{C}$ warped multidimensional signals, for each input sample.
\begin{equation}
\label{eqn1}
\mathcal{W}^{(i,j),\nu} = WARP(\mathfrak{S}^{(i,j)},\mathcal{T}^\nu), \forall i,j,\nu\,
\end{equation}
This warped samples will be used together in the next step, to form the feature vectors.

\subsection{\textbf{Feature Generation and Classification}}
The resulting warped signals of a sample, show the matching of the sample with different templates. We performed the warping with all possible actions, to train our system the response of an input sample when warped with the positive class template and also the negative ones. To consider the localization in both time and frequency domains, we extract features from the warped multidimensional signals by the Wavelet decomposition. The Wavelet decomposition extracts features from the signal with a multilevel algorithm. At each stage, the approximation coefficients and the detail coefficients of the input signal are computed by convolving the signal with a low-pass and a high-pass filter, respectively, followed by decimation blocks. Then the approximation coefficients are fed to the next stage as input. The resulting sets of coefficients represent the low-frequency and high-frequency components of the signal, in different time scales. Here we apply the Wavelet decomposition to the sub-signals of the warped samples. Let $\mathfrak{S}$ be an arbitrary action sample. In the previous step, the warping of $\mathfrak{S}$ with different templates was performed. Suppose $\mathcal{W}^{\nu}$, $\nu = 1,\cdots,\mathcal{C}$ are the resulting warped samples. So, applying the Wavelet decomposition, we will have:
\begin{equation}
\label{eqn2}
\mathcal{F}^\nu_k\ = Wavedec(\mathcal{W}^\nu_k), \forall \nu,k\,
\end{equation}

\begin{figure*}
\includegraphics[width=\linewidth]{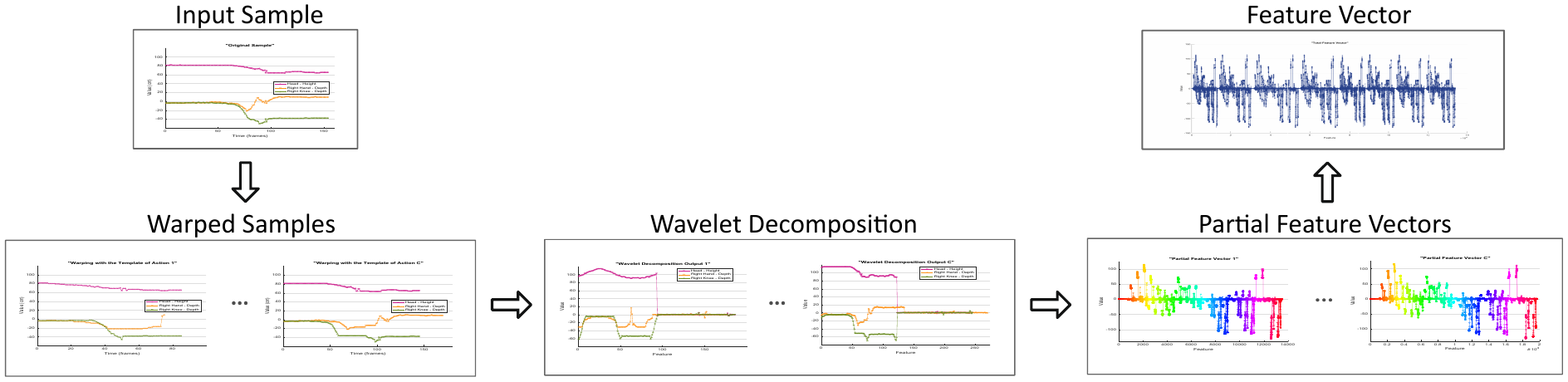}
\caption{An example of the temporal warping and feature vector generation procedures for an arbitrary action sample.} \label{fig:6}
\end{figure*}

The extracted coefficients from the different sub-signals are concatenated to form the feature vector. Since we have warped each specific sample with all of the templates, the extracted features from the warping results, with respect to the different templates, should also be concatenated to each other to form the total feature vector. Note that since we have warped the samples to the action templates previously, the corresponding input signals of the Wavelet decomposition filters have the same length. This causes the filter outputs, and so the total feature vectors to be meaningful for the classification purpose. An example of the temporal warping and feature vector generation algorithms is illustrated in Fig. \ref{fig:6}.

\begin{equation}
\label{eqn3}
\mathcal{F} = (\mathcal{F}^1_1,\cdots,\mathcal{F}^1_\mathcal{K},\cdots,\mathcal{F}^\mathcal{C}_1,\cdots,\mathcal{F}^\mathcal{C}_\mathcal{K})
\end{equation}

The generated feature vectors of the training and testing samples are then used for classification purpose. Here we employ a Random Decision Forest (RDF) classifier. Random forest is an ensemble learning method that fits a number of simple and unpruned decision tree classifiers on various bootstrap samples of the data. Moreover, the split at every node of each tree is made by the best feature from among a random subset of all features. The final prediction is made by the majority vote of all trees in the forest. As each tree makes a high-variance but approximately unbiased prediction, the ensemble of trees reduces the variance and produces a relatively robust and accurate prediction.

\section{\textbf{Experiments}}\label{experiments_section}
The Wavelet decomposition has two parameters: the Wavelet filters type, and the number of levels. In order to choose the appropriate value for this parameters, we perform a parameter tuning procedure within the training data. For this purpose, we divide the training set into two groups. Then we form the feature vectors with the different parameter values and compare the classification results between the groups. The best performing values are used for the original decomposition on the training and testing phases. We search for the best wavelet type and the number of levels between the sets of $\{Daubechies, Coiflet, Symlet\}$ and $\{1,3,5\}$ respectively.

In this section, we evaluate our method on five well-known datasets: Cornell Activity Datasets (CAD-60, CAD-120), UT-Kinect dataset, UCF-Kinect dataset, and TST fall detection dataset. We refer the interested readers for a review on the Kinect activity datasets to \cite{firman2016rgbd} and \cite{zhang2016rgb}. In the following, we will compare the experimental results of our method, with the state-of-the-art skeletal-based methods on each dataset. For some datasets, there may be methods using the depth and RGB modalities, achieving better results. In the cases, that k-fold cross-validation is performed, a random permutation of the subjects is considered. Then the whole process is repeated many times, and the results are averaged.

\subsection{\textbf{CAD-60 Dataset}}
\begin{figure}
\centering
\includegraphics[width=0.75\linewidth]{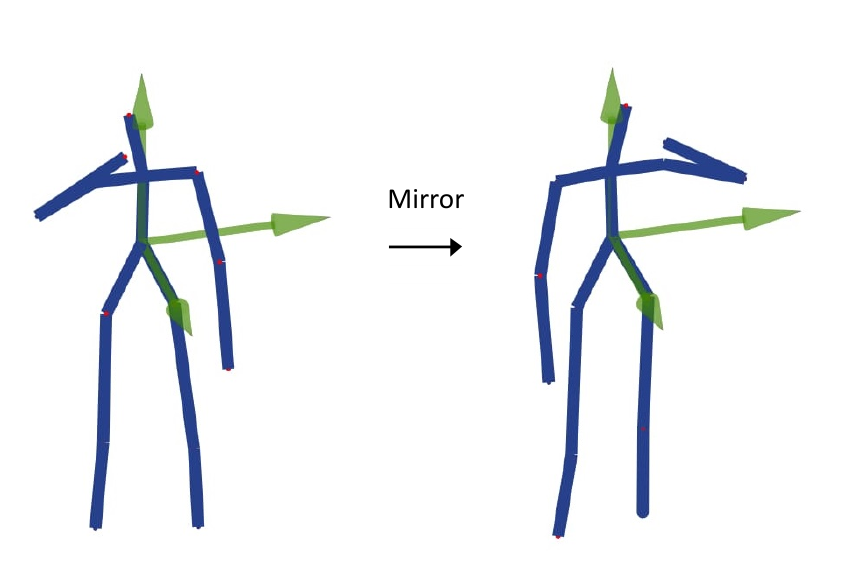}
\caption{An illustration of the skeleton mirroring for the action "Drinking Water" from the \textquotedblleft CAD-60\textquotedblright dataset.} \label{fig:7}
\end{figure}

\begin{figure}
\includegraphics[width=\linewidth]{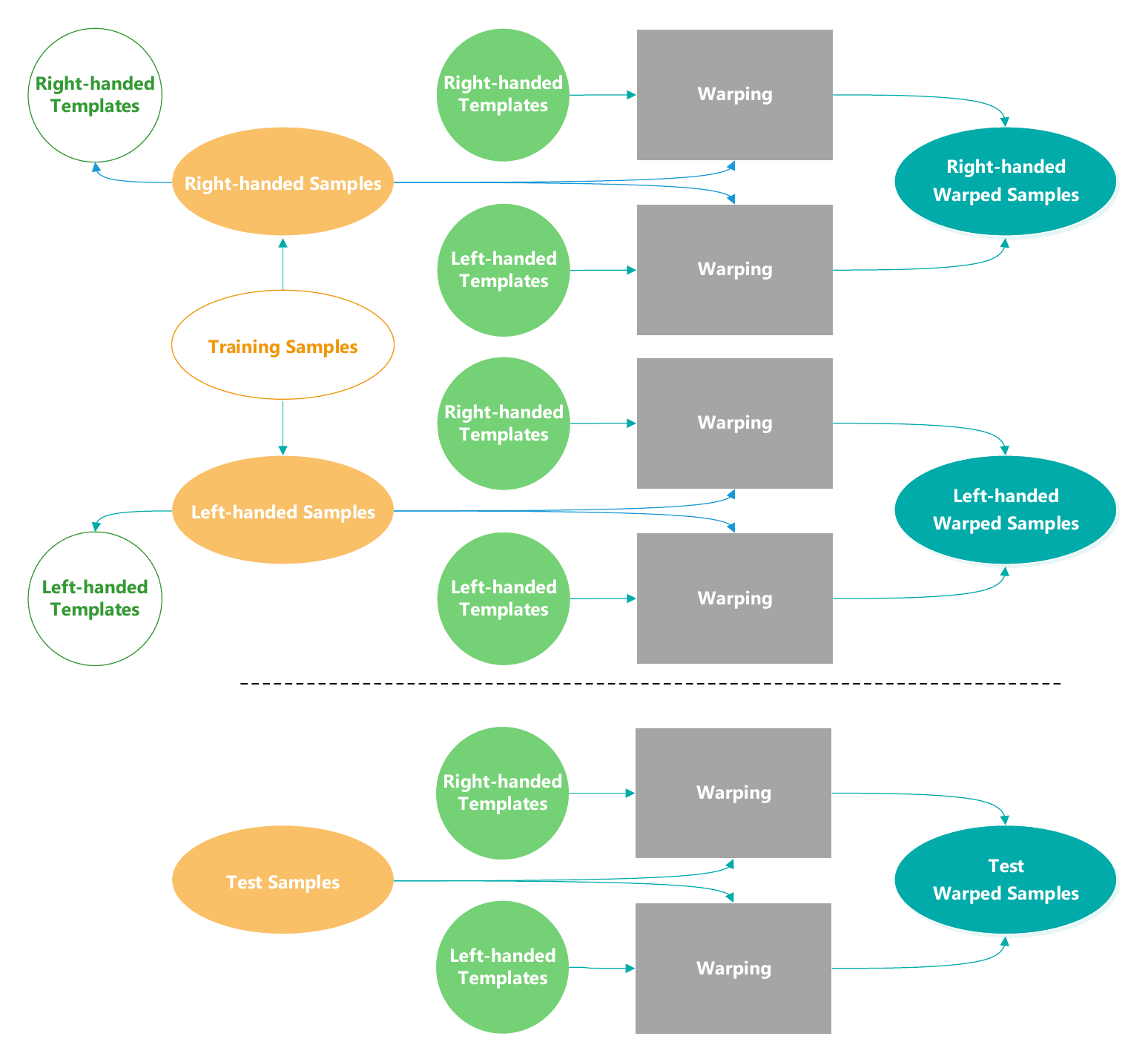}
\caption{Warping procedure, while mirroring the samples.} \label{fig:8}
\end{figure}

The CAD-60 dataset \cite{sung2012unstructured}, is a publicly available dataset captured by the Kinect sensor. In addition to the RGB and depth map modalities, the $3$-D locations of the 15 tracked skeleton joints in each frame are also available in this dataset. It consists of 12 human daily life activities, performed by four subjects in five different environments. The major issue with this dataset is the problem of handedness. Three of the subjects are right-handed, and the other one is left-handed. For example, consider the action of drinking water. Performing this action with the right hand, and with the left hand, will result in quite different joint trajectories, and so they will generate dissimilar feature vectors, while, they belong to the same action class. To address this issue, we adopt the well-known mirroring idea. We create a copy from each action sample in the training set, which is the mirrored version of the original sample along the bisector plane of the body. Therefore, the number of training sample will be twice, while in the test phase, merely the original samples are used. We also create two distinct templates for each action class, one for the left-handed samples and one for the right-handed ones. Then to train our system the response of the samples, to the correct and incorrect warping, we warp each action sample, regardless of its handedness, with both the templates of all classes. The final feature vectors are formed by concatenating the corresponding features of the two templates. Figures \ref{fig:7} and \ref{fig:8} give an illustration of the mirroring and warping procedures respectively.

Following \cite{sung2012unstructured}, we use the same experimental setup. Actions are classified into five environments: office, kitchen, bedroom, bathroom, and living room. Then the Leave One Subject Out (LOSubO) cross-validation is performed for each environment, i.e. three subjects are used for the training, and the test is performed on the other one, for all possible permutations. Table \ref{tbl1} gives the recognition results produced by our method for the different environments. The comparison with the other methods is presented in Table \ref{tbl2}. Except for the recent work by Zhu et al. \cite{zhu2016human}, the recognition results demonstrate that our method is comparable with the state-of-the-arts.

\begin{table}[!t]
\renewcommand{\arraystretch}{1.3}  
\caption{Recognition results on different environments for the \textquotedblleft CAD-60\textquotedblright dataset.}
\label{tbl1}
\centering
\begin{tabular}{l c c}
\hline
\textbf{Environment} & \textbf{Precision} & \textbf{Recall}\\
\hline
Bathroom & 100.0\% & 100.0\%\\
Bedroom & 91.6\% & 93.3\%\\
Kitchen	 & 93.7\% & 95.0\%\\
Living Room & 93.7\% & 95.0\%\\
Office & 87.5\% & 88.7\%\\
\textbf{Average} & \textbf{93.3\%} & \textbf{94.4\%}\\
\hline
\end{tabular}
\end{table}

\begin{table}[!t]
\renewcommand{\arraystretch}{1.3}  
\caption{Comparison of the different methods on the \textquotedblleft CAD-60\textquotedblright dataset.}
\label{tbl2}
\centering
\begin{tabular}{l c c}
\hline
\textbf{Method} & \textbf{Precision} & \textbf{Recall}\\
\hline
Sung et al. \cite{sung2012unstructured} & 67.9\% & 55.5\%\\
Zhu et al. \cite{zhu2014evaluating} & 93.2\% & 84.6\%\\
Faria et al. \cite{faria2014probabilistic} & 91.1\% & 91.9\%\\
Shan and Akella \cite{shan20143d} & 93.8\% & 94.5\%\\
Gaglio et al. \cite{gaglio2015human} & 77.3\% & 76.7\%\\
Parisi et al. \cite{parisi2015self} & 91.9\% & 90.2\%\\
Cippitelli et al. \cite{cippitelli2016human} & 93.9\% & 93.5\%\\
Zhu et al. \cite{zhu2016human} & \textbf{97.4}\% & \textbf{95.8}\%\\
\textbf{our method} & \textbf{93.3\%} & \textbf{94.4\%}\\
\hline
\end{tabular}
\end{table}

\subsection{\textbf{CAD-120 Dataset}}
The CAD-120 dataset \cite{koppula2013learning}, is originally a high-level human activity dataset. It includes ten complex activities, performed by four subjects for three times. Each action consists of a sequence of atomic activities called sub-activities. Our motivation to choose the CAD-120 dataset was the importance of the object manipulations in the activities of this dataset. All of the ten high-level activities include human object interactions. In some cases, e.g. the stacking objects and unstacking objects, the discrimination between the actions is significantly caused by the objects. In this dataset, an object tracking algorithm was applied on the RGB images of the frames of all the samples, and the 2D locations of the objects bounding boxes were specified. We have used the bounding boxes to extract the $3$-D location of the objects using the corresponding depth map images.

\begin{figure}
\centering
\includegraphics[width=0.9\linewidth]{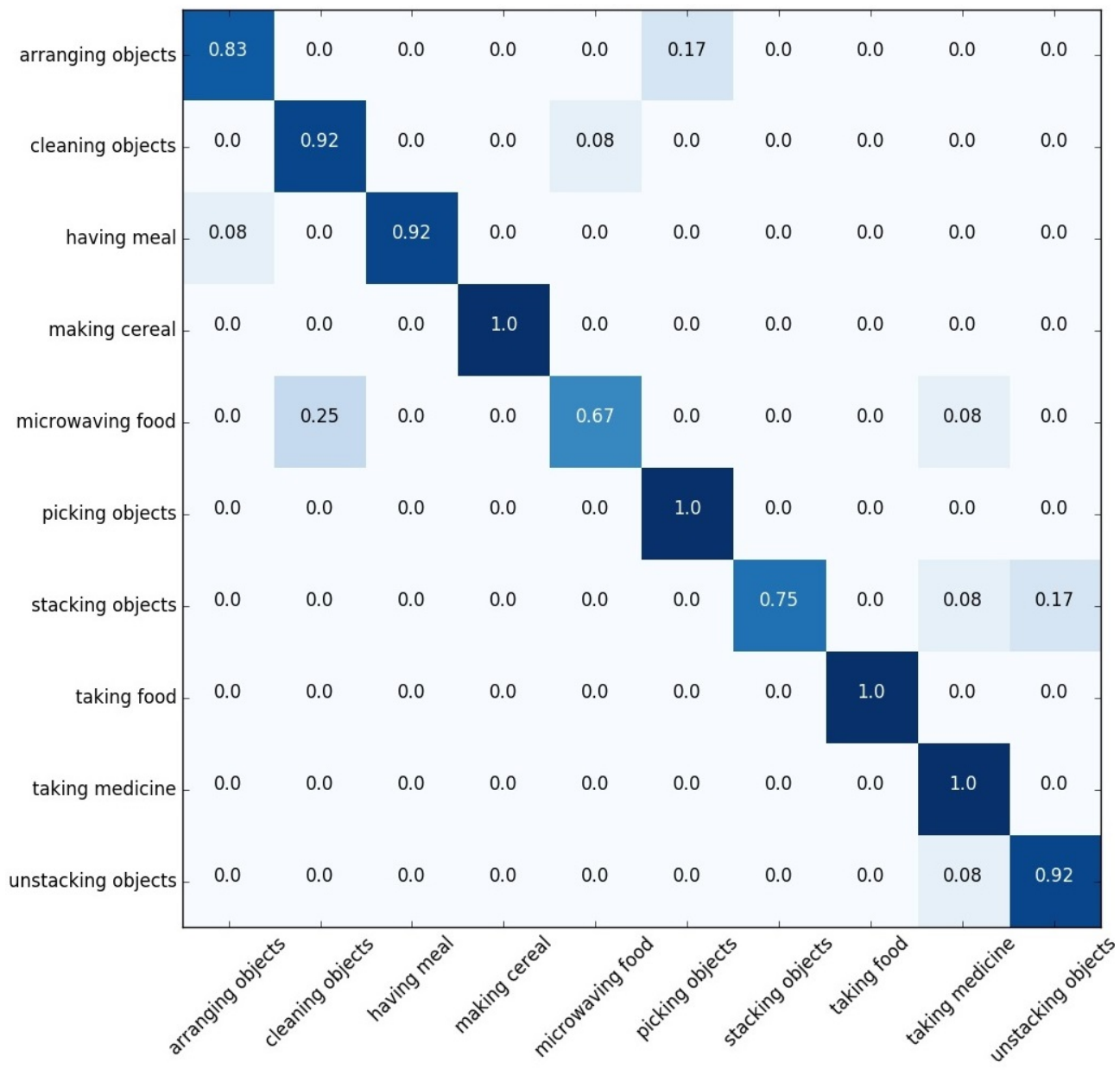}
\caption{Confusion matrix for the \textquotedblleft CAD-120\textquotedblright dataset.} \label{fig:9}
\end{figure}

Although our method does not concentrate on the high-level activities, the evaluation results on this dataset demonstrate comparable performance of our method with the state-of-the-arts. The confusion matrix is presented in Fig. \ref{fig:9}. As this figure shows, the main trouble with this dataset is about confusing the activities \textquotedblleft stacking objects\textquotedblright with \textquotedblleft unstacking objects\textquotedblright, \textquotedblleft microwaving food\textquotedblright with \textquotedblleft cleaning objects\textquotedblright, and \textquotedblleft arranging objects\textquotedblright with \textquotedblleft picking objects\textquotedblright, which are very similar. Comparison of our method with the state-of-the-arts is shown in Table \ref{tbl3}. In the dataset, the ground-truth temporal segmentation of the actions was provided. Some hierarchical methods have used this segmentation data to improve their results. Since our method recognizes the high-level actions in one stage, we have not used this data.

\begin{table}[!t]
\renewcommand{\arraystretch}{1.3}  
\caption{Comparison of the high-level recognition accuracies of the different methods on the \textquotedblleft CAD-120\textquotedblright dataset.}
\label{tbl3}
\centering
\begin{tabular}{l c c}
\hline
\textbf{Method} & \textbf{Without ground-truth} & \textbf{With ground-truth}\\
\hline
Koppula et al. \cite{koppula2013learning} & 80.6\% & 84.7\%\\
Hu et al. \cite{hu2014learning} & 87.0\% & -\\
Tayyub et al. \cite{tayyub2014qualitative} & \textbf{95.2\%} & -\\
Taha et al. \cite{taha2015skeleton} & - & 94.4\%\\
Koppula and Saxena \cite{koppula2016anticipating} & 83.1\% & 93.5\%\\
\textbf{our method} & \textbf{90.1\%} & -\\
\hline
\end{tabular}
\end{table}

\subsection{\textbf{UT-Kinect Dataset}}
The UT-Kinect dataset was introduced in \cite{xia2012view}. The dataset consists of ten actions: walk, sit down, stand up, pick up, carry, throw, push, pull, wave and clap hands. Each action is performed twice by ten different subjects in a lab environment, and 20 skeleton joints are tracked in each frame. The relatively high within-class variance is a considerable challenge with this dataset. The different actions of this dataset are performed continuously by each subject, and the temporal segmentation is manually provided.

To be comparable with the previous works, we have tested our algorithm using 2-fold cross subject validation setting, i.e. for a random permutation of the subjects, half of them were used for the training and the remaining for testing, and then vice versa.  The comparison of our method with the state-of-the-arts is presented in Table \ref{tbl4}. It should be mentioned that Xia et al. \cite{xia2012view}, and Cippitelli et al. \cite{cippitelli2016human} had reported 90\%, and 95.1\% recognition accuracies respectively, using the Leave One Sequence Out (LOSeqO) experimental setup. Also, Liu et al. \cite{liu20163d} and Yang et al. \cite{yang2016latent} had achieved the 95.5\% and 98.8\% accuracies, adopting the Leave One Subject Out (LOSubO) and 10-fold cross-validation settings, respectively. Since these experimental settings are rather easier in comparison with the 2-fold method, we have reported in Table \ref{tbl4} only the methods which have adopted the 2-fold setting.

\begin{table}[!t]
\renewcommand{\arraystretch}{1.3}  
\caption{Comparison of the different methods on the \textquotedblleft UT-Kinect\textquotedblright dataset, using the Cross Subject setting.}
\label{tbl4}
\centering
\begin{tabular}{l c}
\hline
\textbf{Method} & \textbf{Accuracy}\\
\hline
Vemulapalli et al. \cite{vemulapalli2014human} & \textbf{97.0\%}\\
Antunes et al. \cite{antunes2016revisit} & 95.1\%\\
Gupta and Bhavsar \cite{gupta2016scale} & 96.0\%\\
\textbf{our method} & \textbf{96.8\%}\\
\hline
\end{tabular}
\end{table}

\subsection{\textbf{UCF-Kinect Dataset}}
Ellis et al. \cite{ellis2013exploring} presented the UCF-Kinect dataset to evaluate their latency-aware learning algorithm, which focuses on reducing the recognition latency. The dataset was captured using a Kinect sensor with the OpenNI platform, which provides the $3$-D coordinates of the 15 skeleton joints. It contains 16 short actions, performed by 16 subjects for five times. Similar to the experimental setting in \cite{ellis2013exploring}, we use the 4-fold cross subject validation as evaluation protocol for this dataset. The comparison with the other methods is shown in Table \ref{tbl5}. Slama et al. \cite{slama2015accurate} reported the 97.9\% recognition accuracy, for a 0.7 and 0.3 split on the 1280 samples of the dataset, for the training and testing sets. Also, Jiang et al. \cite{jiang2013robust} had achieved the 98.7\% accuracy, adopting the 2-fold setting on the samples.

\begin{table}[!t]
\renewcommand{\arraystretch}{1.3}  
\caption{Comparison of the different methods on the \textquotedblleft UCF-Kinect\textquotedblright dataset.}
\label{tbl5}
\centering
\begin{tabular}{l c}
\hline
\textbf{Method} & \textbf{Accuracy}\\
\hline
Zanfir et al. \cite{zanfir2013moving} & 98.5\%\\
Kerola et al. \cite{kerola2014spectral} & 98.8\%\\
Yang et al. \cite{yang2014effective} & 97.1\%\\
Beh et al. \cite{beh2014hidden} & \textbf{98.9}\%\\
Ding et al. \cite{ding2015stfc} & 98.0\%\\
Lu et al. \cite{lu2016efficient} & 97.6\%\\
\textbf{our method} & \textbf{97.9\%}\\
\hline
\end{tabular}
\end{table}

\subsection{\textbf{TST Fall Detection Dataset}}
\begin{figure}
\centering
\includegraphics[width=0.9\linewidth]{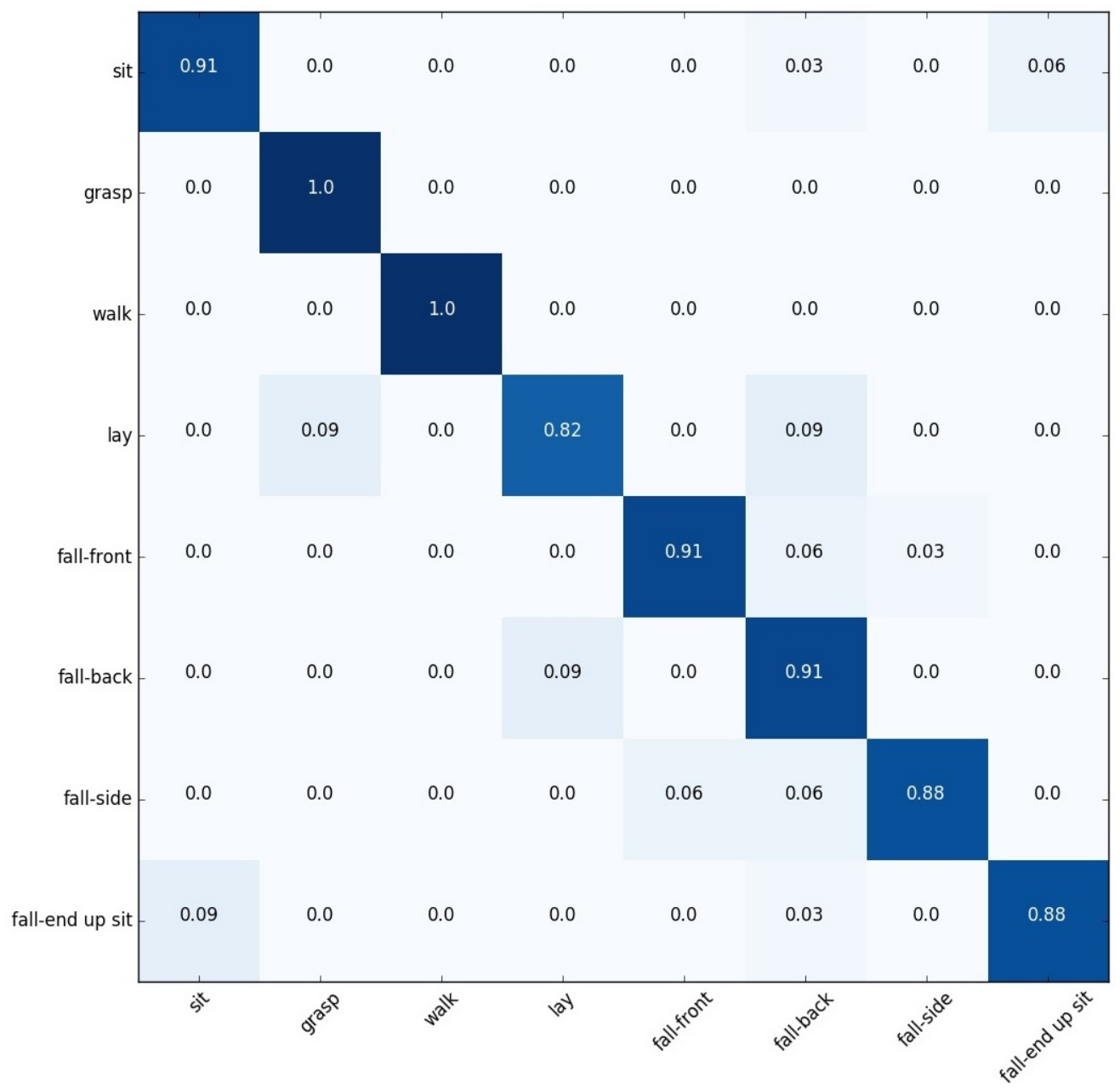}
\caption{Confusion matrix for the \textquotedblleft TST Fall Detection\textquotedblright dataset.} \label{fig:10}
\end{figure}

This dataset was originally collected by Gasparrini et al. \cite{gasparrini2016proposal} as a part of a study on the human fall event detection problem. They aimed at using the fusion of camera and wearable sensors to detect the fall event. The dataset was collected using the Microsoft Kinect v2 and the Inertial Measurement Unit (IMU) sensors. In this dataset two groups consisting of four daily living actions and four fall actions were performed by 11 subjects for three times. Although the wearable sensors provide very valuable data, we don't use this modality in our work and perform the recognition just utilizing the tracked skeleton joints data. Same as \cite{gasparrini2016proposal}, we evaluated our method with the Leave One Subject Out cross-validation (LOSubO) setting. The average accuracy of our method for all the activities is 92.8\%. Note that in \cite{gasparrini2016proposal} the 99\% recognition accuracy is reported using the multiple modalities, including the wearable sensors, and so the results are not comparable. The confusion matrix of our method is illustrated in Fig. \ref{fig:10}.

\section{\textbf{Conclusion}}\label{conclusion_section}
In this paper, we have developed a trajectory-based activity recognition system. We represented a human action as a set of time series corresponding to the normalized coordinates of the skeleton joints. Our representation is also able to simultaneously model the interaction between human and objects in the scene. Then we introduced an algorithm to effectively construct templates for joint and object trajectories. Also, a DTW-based warping procedure was proposed to alleviate the effects of variations in the styles of performing actions. The wavelet filters were utilized to extract meaningful features from the signals, and the classification was performed by the Random Decision Forests. The experimental evaluation of the proposed method on several public datasets yielded comparable performance to the state-of-the-arts. Although our proposed method works well on the recognition of simple and short actions, the template-based approaches have problems with the more complex activities. Representing the activities which consist of multiple simple sub-actions using one unique template, will not have good recognition results, due to their nature. So next we plan to apply modifications to our method to make it usable for the complex human activities. 

\section*{\textbf{Acknowledgment}}\label{acknowledgment}
This work was supported by a grant from Iran National Science Foundation (INSF).

\section*{\textbf{References}}\label{references}

\bibliography{references}
\bibliographystyle{elsarticle-num}

\end{document}